\documentclass{article}

\usepackage{arxiv}

\usepackage[utf8]{inputenc} 
\usepackage[T1]{fontenc}    
\usepackage{hyperref}       
\usepackage{url}            
\usepackage{booktabs}       
\usepackage{amsfonts}       
\usepackage{nicefrac}       
\usepackage{microtype}      
\usepackage{lipsum}		
\usepackage{graphicx}
\usepackage{natbib}
\usepackage{doi}

\title{Fighting Fires from Space: Leveraging Vision Transformers for Enhanced Wildfire Detection and Characterization}

\date{December 8, 2022}	

\author{ \href{https://orcid.org/0000-0002-4406-9937}{\includegraphics[scale=0.06]{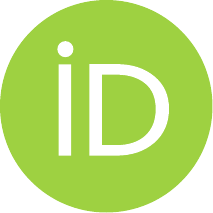}\hspace{1mm}Aman Agarwal}\thanks{This research was conducted when the authors were at Indiana University, Bloomington, IN 47405.} \\
	Department of Informatics, \\ 
    Computing, \& Engineering \\
	Bloomington, IN 47405 \\
	\texttt{amanagar@iu.edu} \\
	\And
	James Gearon \\
	Department of Earth \& \\ Atmospheric Sciences \\
	Bloomington, IN 47405 \\
	\texttt{jhgearon@iu.edu} \\
    \And
	Raksha Rank \\
	Department of Informatics, \\ 
    Computing, \& Engineering \\
	Bloomington, IN 47405 \\
	\texttt{rrank@iu.edu} \\
    \And
	Etienne Chenevert \\
	Department of Earth \& \\ Atmospheric Sciences \\
	Bloomington, IN 47405 \\
	\texttt{etachen@iu.edu} \\
}

\renewcommand{\shorttitle}{Agarwal et al.}

\hypersetup{
pdftitle={A template for the arxiv style},
pdfsubject={q-bio.NC, q-bio.QM},
pdfauthor={David S.~Hippocampus, Elias D.~Striatum},
pdfkeywords={First keyword, Second keyword, More},
}

\begin{document}
\maketitle

\begin{abstract}
	Wildfires are increasing in intensity, frequency, and duration across large parts of the world as a result of anthropogenic climate change. Modern hazard detection and response systems that deal with wildfires are under-equipped for sustained wildfire seasons. Recent work has proved automated wildfire detection using Convolutional Neural Networks (CNNs) trained on satellite imagery are capable of high-accuracy results. However, CNNs are computationally expensive to train and only incorporate local image context. Recently, Vision Transformers (ViTs) have gained popularity for their efficient training and their ability to include both local and global contextual information. In this work, we show that ViT can outperform well-trained and specialized CNNs to detect wildfires on a previously published dataset of LandSat-8 imagery \cite{Pereira2021}. One of our ViTs outperforms the baseline CNN comparison by 0.92\%. However, we find our own implementation of CNN-based UNet to perform best in every category, showing their sustained utility in image tasks. Overall, ViTs are comparably capable in detecting wildfires as CNNs, though well-tuned CNNs are still the best technique for detecting wildfire with our UNet providing an IoU of 93.58\%, better than the baseline UNet by some 4.58\%. 
\end{abstract}

\keywords{CNN \and Transformers \and Wildfire}

\begin{figure}
	\centering
	\includegraphics[width=0.7\linewidth]{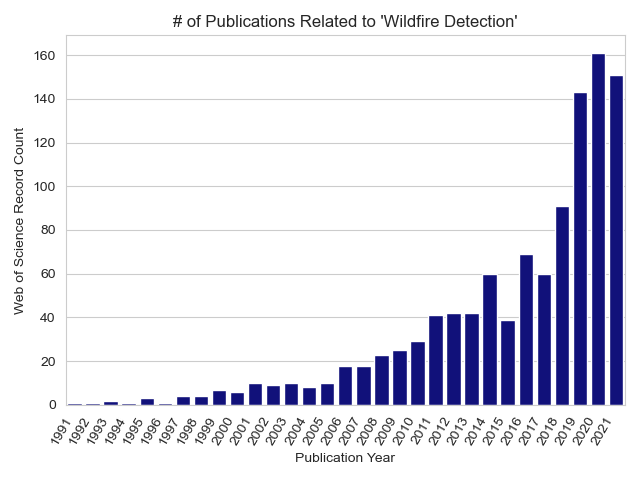}
	\caption{Publication counts containing the keyword ``Wildfire Detection'' in recent years \it{source: Web of Science}.}
	\label{fig:publications}
\end{figure}

\section{Introduction}
\label{sec:intro}

As a consequence of anthropogenic climate perturbation along with a century of fire-management practices that left considerable amounts of natural fuel covering large parts of the land-surface, wildfires have increased in frequency, intensity, and duration over the past two decades \cite{UN2022}. Wildfires can have disastrous impacts on ecosystems, human society, and the global economy. Indeed, changes in air-quality in the western US, Australia, and elsewhere, have increased in recent years leading to healthcare impacts (IPCC). Furthermore, wildfire has ravaged agricultural and pastoral land, leading to quantifiable economic losses. It is therefore imperative to detect wildfire activity as early as possible so that real-time risking can be done and relevant information relayed to regional and local officials. The conventional method for early fire detection consists of specialized fire lookouts or localized camera monitors deployed in previously identified fire-prone areas \cite{Dewangan2022}. Clearly, these methods are only effective at a local scale, whereas wildfire is a regional or landscape-scale phenomenon. The era of satellite observation has allowed for complete near-real time coverage of the global landscape. In this work, we aim to recreate a published method \cite{Pereira2021} for wildfire detection using Vision Transformers (ViT) trained on a suite of labeled satellite images. These techniques that incorporate global imagery can scale to match the spatial area of regional wildfire events and can be deployed to give updates.

The utility of an accurate, robust, and efficient wildfire detection system cannot be understated. Wildfire studies is a fast-growing subfield of earth science that has received increased attention from both state and federal government funding agencies over the last 30 years (Figure \ref{fig:publications}). Record counts from the popular Web of Science database demonstrate the increased demand for studies concerning automated or semi-automated wildfire detection. An additional motivation for this work is the need in the remote sensing community for computer vision practitioners and subject matter experts. Both fields involve image processing, but rarely communicate. Furthermore, the amount of data produced and telemetered by earth observing satellites is immense, amounting to approximately 10 Terabits per day and growing still \cite{UN2022}. This volume and velocity of data necessitates sophisticated computer vision algorithms for processing. However, this volume of data also makes training remote-sensing based algorithms like convolutional neural networks computationally costly, as individual images can be several megabytes in size and contain 10s of channels in the form of image bands. To realize efficient and reliable remote wildfire detection, it is imperative for the designer to reduce training time as much as possible, as to incorporate new data quickly.

\section{Background}
Recently, `Vision Transformers' or ViTs have become popular in the field of Computer Vision for their efficiency and flexibility for a suite of vision related tasks. ViTs require substantially less computational resources, making them a fit tool to identify wildfires remotely and accelerate emergency response time. The ViT concept comes from a pioneering publication by Dosovitskiy et al. \cite{dosovitskiy2020image}, where the authors implement a transformer architecture for images and compare their results against the a CNN architecture, finding that they are able to achieve similar results as the CNN by subdividing the input image into `patches' and applying a transformer to the patches as they propagate information through the network. As applied to remote sensing, the ViT architecture has recently been used by Bazi et al. \cite{bazi2021vision} where they test the model’s performance in land type classification. This is a promising finding that indicates that the ViT architectural approach can be readily applied to many scientific questions in remote sensing as long as there is sufficient training data and a desired classification output. 

\begin{figure*}
	\centering
	\includegraphics[width=\linewidth]{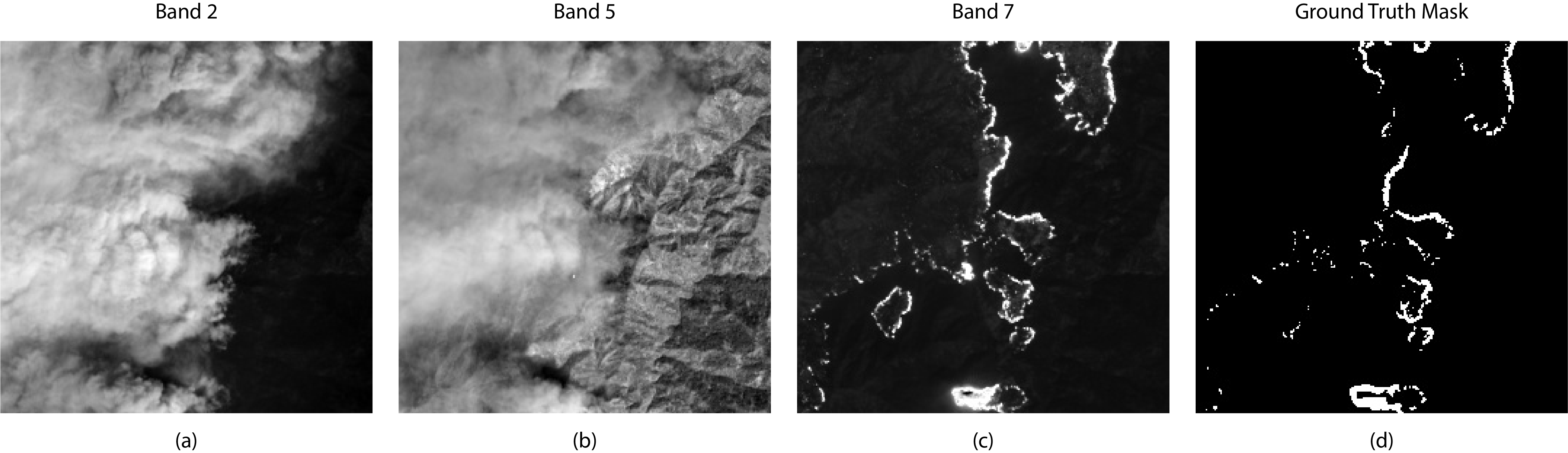}
	\caption{Fire as visualized from three different bands of a Landsat image. Part (a), (b), and (c) show a range of bands (Blue, Near Infrared, and SWIR2) to demonstrate transparency to clouds at different wavelengths, and part (d) shows the ground truth mask of wildfire. Bands 2 (Blue), 3 (Green), and 4 (Red) fall in the visible light spectrum and may not show fire most of the times due to occlusion by clouds and smoke. This is also the case for the near infrared portion of the spectrum (band 5 NIR). However, band 6 (SWIR1) and 7 (SWIR2) part of the short wave infrared spectrum and are therefore resolve fire more readily.}
	\label{fig:band_compare}
\end{figure*}

In this work, we propose to integrate the Vision Transformer introduced by Dosovitskiy et al. \cite{dosovitskiy2020image} into the wildfire detection framework of Pereira et al. \cite{Pereira2021}. As a part of that project, the authors generated a large (200 GB compressed) dataset of annotated LandSat-8 Operational Land Imager (OLI) satellite images for active fire detection using semantic segmentation. More specifically, the authors include a comparison of handcrafted fire detection algorithms using Landsat spectral bandwidth data and three variations of CNN algorithms derived from a U-Net architecture. The first network (U-Net 10c) was assembled with the traditional U-Net structure taking a 10-channel Landsat image as input, the second design (U-Net 3c) sought to reduce memory usage by taking a 3-channel Landsat image as input, and a third design (U-Net-Light 3c) was created from a structure like U-Net 3c, but with the number of filters per layer being reduced by 4 times. The results proved that the author's trained CNNs were able to reproduce the fire detection accuracy of previous fire detection algorithms reasonably well. Interestingly, they found each of the CNN model designs performed to roughly the same accuracy, indicating that large, complex models (U-Net 10c) are not necessary in properly identifying wildfires. Rather, smaller, less complex models (U-Net 3c and U-Net-Light 3c) created from 3 channel inputs, specifically channels 7, 6, and 2 from the Landsat spectral data. They highlight that because this was a pioneering study on the implementation of CNN’s for fire detection, future lines of research could seek to optimize the performance of CNN’s on wildfire detection via implementation of ensemble neural networks, implementation of spatial and temporal boundaries on fire detection (to exclude non-wildfire fires), and image resolution enhancement. We argue the inclusion of ViTs into a fire detection workflow represents a step forward in the optimization of effective fire warning models.

Similar work has been done by Rashkovetsky et al. \cite{Rashkovetsky2021} on Sentinel-2 Multi Spectral Instrument (MSI) images in which the authors combine different sensor data from multiple sources in visible, infrared and microwave domains. The authors created a workflow where they investigate four different sources of satellite data: C-Band Synthetic Aperture Radar (C-SAR), multi-spectral imagery, sea and land surface temperature, and hyperspectral MODIS imagery on different Sentinel-2 MSI images. Furthermore, they use a U-Net based CNN architecture to train on the generated dataset. The authors take into consideration the images in the presence of clouds as well as clear conditions for all the four sources. Output evaluation was done for the images that were acquired in the time frame of 12h in the same scene. Further, for testing purposes, 60 images were selected from all the 4 instruments in the time frame of 12h. As pixels that contain information regarding fire would be in small portions of the scene, this could eventually lead to convergence of the model in the local minima. To tackle this problem, they try two different loss functions,  binary cross entropy (BCE) loss and Dice loss. Post various studies and experimentation they find that the Sentinel-1 and Sentinel-2 instruments work best for cloudy conditions and Sentinel-2 and Sentinel-3 instruments show better results in clear conditions.

The work by Barmpoutis et al. \cite{Barmpoutis2020} relays multiple approaches for smoke and fire detection systems; namely terrestrial, aerial, and satellite-based approaches. Terrestrial systems are more accurate and efficient as they can work closely with the affected area, however, the proximity makes it difficult for them to cover a large area. Aerial systems like Unmanned Aerial Vehicles (UAV) provide much larger area coverage but suffer from high cost of operation. On the other hand, satellite based detection has a huge coverage but there can be delays in data acquisition of hours to days.

The application of computer vision to wildfire detection extends beyond the satellite domain. terrestrial detection systems have gained popularity in recent years. Dewagan et al. \cite{Dewangan2022} proposed a solution to detect fires in real time from video feeds in specific locations. The authors focus on detecting forest fires with the help of smoke. They also created a dataset of labeled wildfire smoke images from cameras in Southern California. The dataset was named FIgLib and the network was named SmokeyNet. At present, independent fire lookout expert teams are responsible for spotting fires based on camera feeds and manual monitoring. But, that does not scale well and moreover, it is slow and prone to error because of manual work required. SmokeyNet is an example of an innovative architecture which combines multiple small networks for different classification tasks. Specifically, a pre-trained ResNet is used to extract image features, these image features are passed to an LSTM network to extract temporal information from two consecutive frames. And the output from these LSTM goes to a Transformer network that does the final classification. They do various ablation studies and find that the combination of small models outperforms the standard CNN-based networks. However, the authors report a high quantity of false positives due to clouds and haze inherent in the RGB camera images.

\section{Methods \& Experimentation}
\subsection{Dataset}
The project is based upon the work of Pereira et al. 2021 \cite{Pereira2021} in which the authors have generated a large (200GB compressed) dataset of annotated LandSat-8 Operational Land Imager (OLI) satellite images for active fire detection using semantic segmentation. We use this annotated dataset for our project. Satellite imagery is abundant, and automated labelling is preferred due to the volume of the data. To this end, three separate mathematical models are used to generate masks: Schroeder et al. \cite{schroeder2016active}, Murphy et al. \cite{murphy2016hotmap}, and Kumar and Roy \cite{kumar2018global}. These models are essentially fit-for-purpose classification algorithms that intake spectral information to decide on a mask. In addition, there is a separate manually annotated test data (9044 images). We use the algorithmically generated data masks for training and validating the models and keep the manually annotated data exclusively for testing.

LandSat-8 OLI images are encoded in TIFF format with a resolution of $ \approx 7600 \times 7600 $ pixels and 16 bits information. There are a total of 11 bands in each image, excluding the panchromatic channel which is a black and white high resolution band for pan-sharpening processes. The ground truth mask consists of two more categories formed by the combination of \cite{schroeder2016active}, \cite{murphy2016hotmap}, and \cite{kumar2018global}: intersection and voting. Intersection produces a mask for pixels that are detected as fire in all of the above algorithms, while voting produces a mask wherever more than one algorithm agrees on a pixel. A total of around 150,000 labeled images of re-scaled size 256x256 are formed using this dataset which are sufficiently large enough to train a deep neural network.

Different bands of the multi-spectral satellite images portray different information. Some bands are also task-specific. For example, Figure \ref{fig:band_compare} shows three different bands along with the wildfire ground truth mask. It can be clearly observed that the bands near visible light in part (a) and (b) are incapable of showing any signs of fire due to occlusion by cloud and smoke. However, bands near infrared (c) are better suited for fire detection tasks. This fact is also supported by the work of Pereira et al. \cite{Pereira2021} in which they obtained comparable results on separate models trained with only three channels (band 7, 6, and 2) and with all the 10 channels.


\subsection{Models}

Transformer networks, which are now a de-facto standard in Natural language processing applications, were the inspiration for Vision Transformers (ViTs). ViTs have become popular as they are simple to parallelize on GPU hardware, making them quick to train and easy to scale to billions of parameters. Furthermore, unlike CNNs, ViT's multi-head self-attention mechanism allows them to observe a global context rather than being limited to only a local region. Recently, architectures like DeiT \cite{touvron2021training}, Swin \cite{liu2021swin}, and CvT \cite{wu2021cvt} have demonstrated that with adequate pre-training and data augmentation, ViTs can not only compete, but surpass the performance of CNN-based networks in image classification tasks. In this work, we compare the performance of two ViT models (Swin-Unet and TransUNet) and one UNet implementation using CNN. We additionally attempted to use the Mask2Former model with poor success (Appendix) \cite{chang21maskformer}.

\begin{figure}
	\centering
	\includegraphics[width=0.5\linewidth]{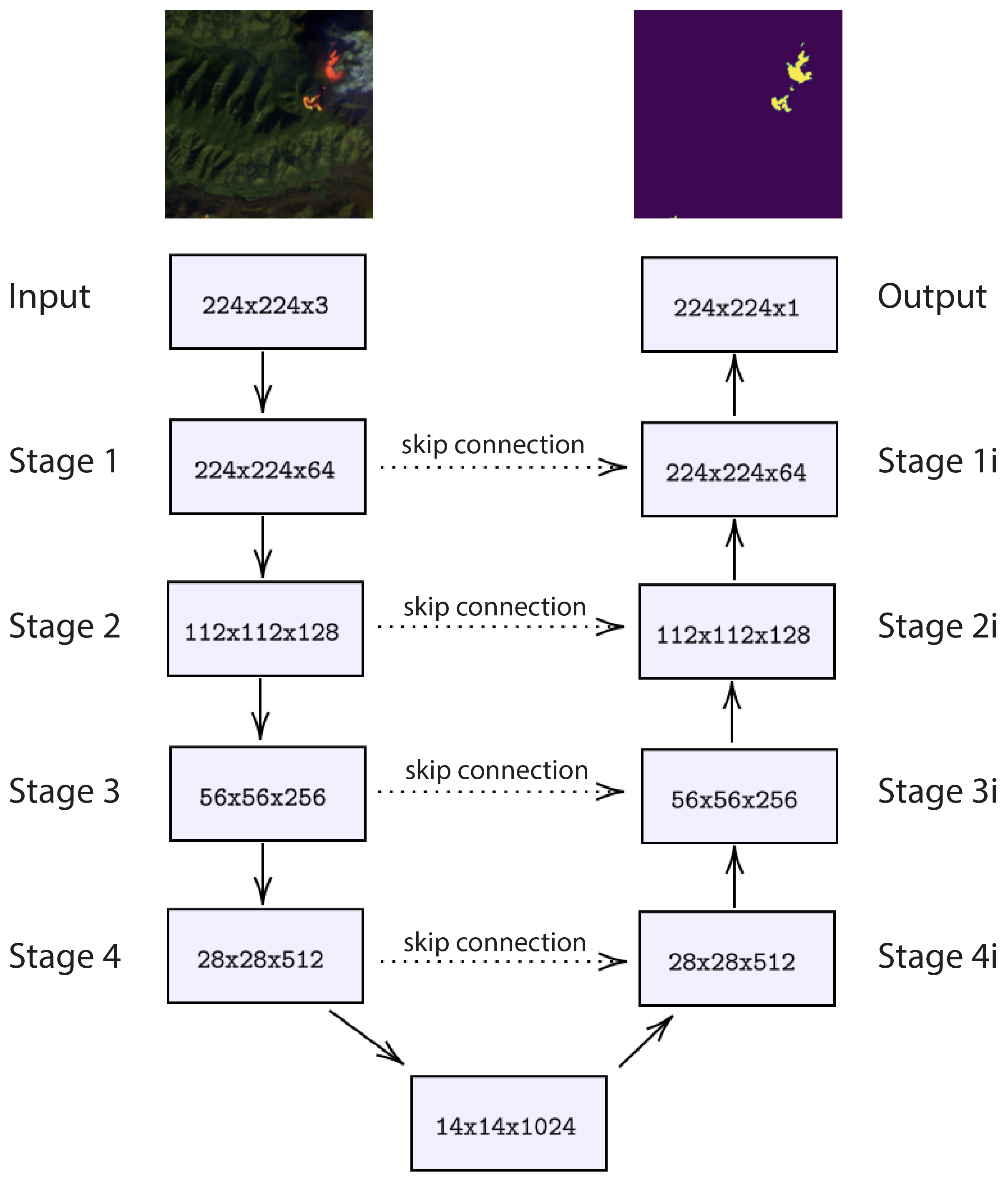}
	\caption{A high level representation of our UNet model architecture. Each transition stage applied two dilated convolutions, batch normalization, and ReLU layers sequentially, followed by strided convolution to reduce the spatial dimension and increase the feature channels.}
	\label{fig:unet}
\end{figure}

\subsubsection{UNet}
The UNet is a classic segmentation model for images pioneered by Ronneberger et al. \cite{ronneberger2015u} for medical image tasks. We use our own implementation of the UNet model on PyTorch, which uses strided convolutions instead of pooling layers for down sampling and dilated convolutions. The network architecture is shown in Figure \ref{fig:unet}. 

Our convolution block consisted of two subsequent sub-blocks of dilated convolution, batch normalization, and ReLU activation functions applied sequentially. Down sampling was done using strided convolution with a patch size of 3 and stride size of 2 between each stage of the UNet. During each down sampling, the feature dimensions were also doubled. We employed transpose convolutions to up sample the feature representation once we reached the bottleneck. There were four stages in our network, the dilation rates for convolution layers in each stage were as follows: 1, 1, 2, 3. The same dilations were used for up sampling as well. There were skip connections between stages of encoders and decoders, much like in the regular UNet.

\subsubsection{TransUNet}
Due to the limitations of CNN to see the global context, Chen et al. \cite{chen2021transunet} fused ViT encoder architecture and CNN to form a UNet. This brought together the detailed high-resolution spatial information from CNN and global context from Transformer's self-attention. We used the Transformer implementation by Adaloglou et al. \cite{adaloglou2021transformer} with 8 ViT blocks, and 512 embedding dimensions.
 
\subsubsection{Swin-Unet}
Liu et al. \cite{liu2021swin} proposed the Swin Transformers that used shifted windows to compute self-attention on image patches and followed a hierarchical (with multiple stages) Transformer architecture. This concept was borrowed by Cao et al. \cite{cao2021swin} for image segmentation task. They used shifted windows concept to form a UNet, and achieved great performance improvement on medical image segmentation task. Swin-Unet contains a Transformer encoder and a decoder, connected via skip connections. We used the default, tiny architecture with patch size of 4 and 96 channels in the first stage.

\begin{figure}
	\centering
	\includegraphics[width=\linewidth]{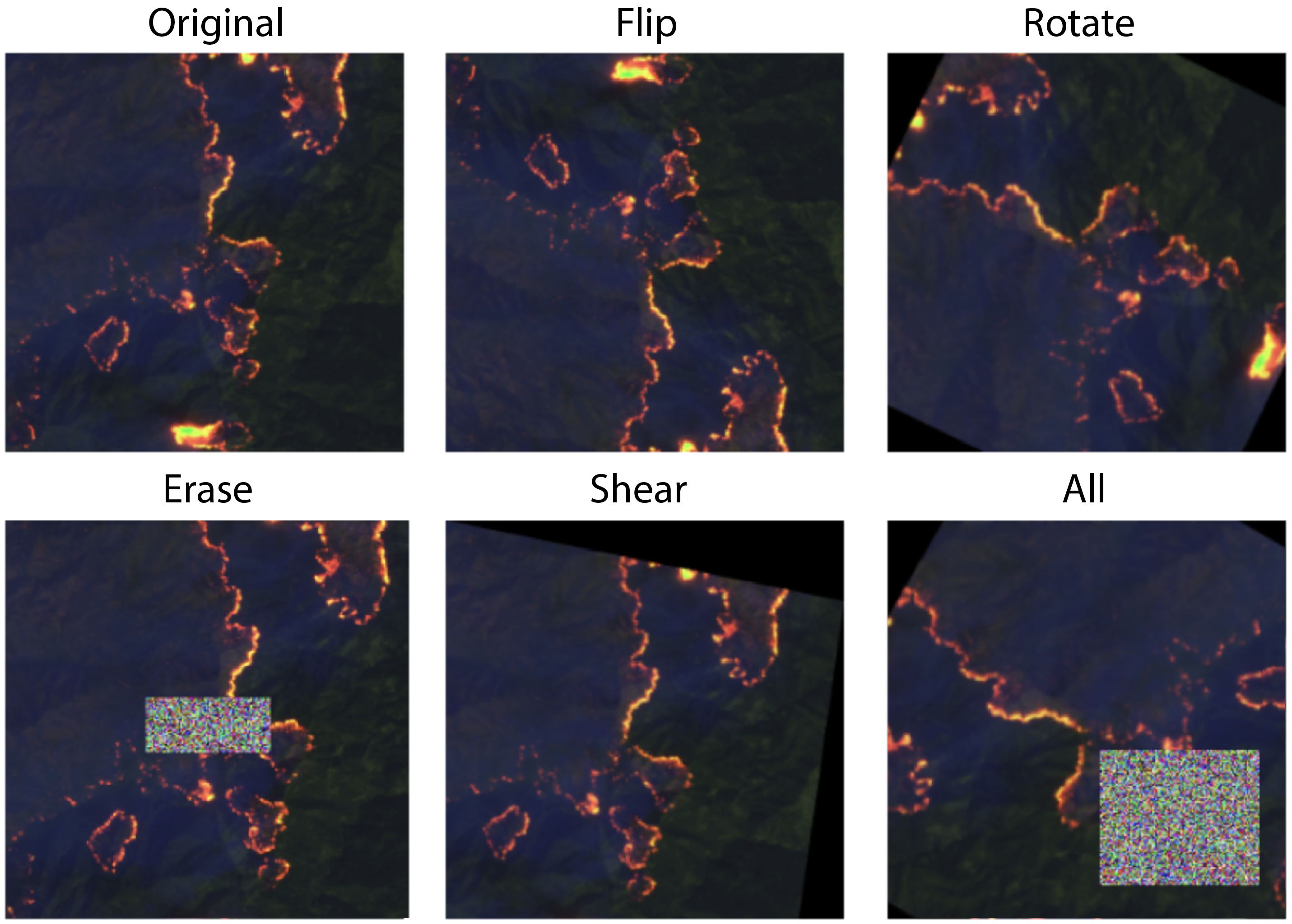}
	\caption{Various data augmentation methods applied to the input images and masks. Each transformation was applied randomly with a certain probability value.}
	\label{fig:augmentation}
\end{figure}

\subsection{Training}
We used PyTorch to train all of our models for 30-50 epochs on an NVIDIA V100 GPU system for 10 hours. One-cycle learning rate scheduler with cosine annealing strategy was used to vary the learning rate while training, with the initial rate of $1e-3$. AdamW was used as the optimizer to tune the network weights with a batch size of $64$ for all the models. Dice Similarity Coefficient (DSC) or dice loss was chosen as the loss function.

Random flipping, rotation, shear, and erase were utilized as standard data augmentation techniques. In random erase, a tiny patch of the image was picked and replaced with noise; the associated mask in that patch was replaced with zero values. All of the image transformations were done at random with a particular probability value: flipping and rotation was done 50\% of the time, shear and erase was done 20\% of the time. Following the transformations, a 224x224 random patch was extracted and sent for training. During validation, the center crop was extracted. Figure \ref{fig:augmentation} shows the data augmentation methods used for training. The code for model training and evaluation is available at \url{https://github.com/amanbasu/wildfire-detection}.

\begin{figure*}
	\centering
	\includegraphics[width=0.9\linewidth]{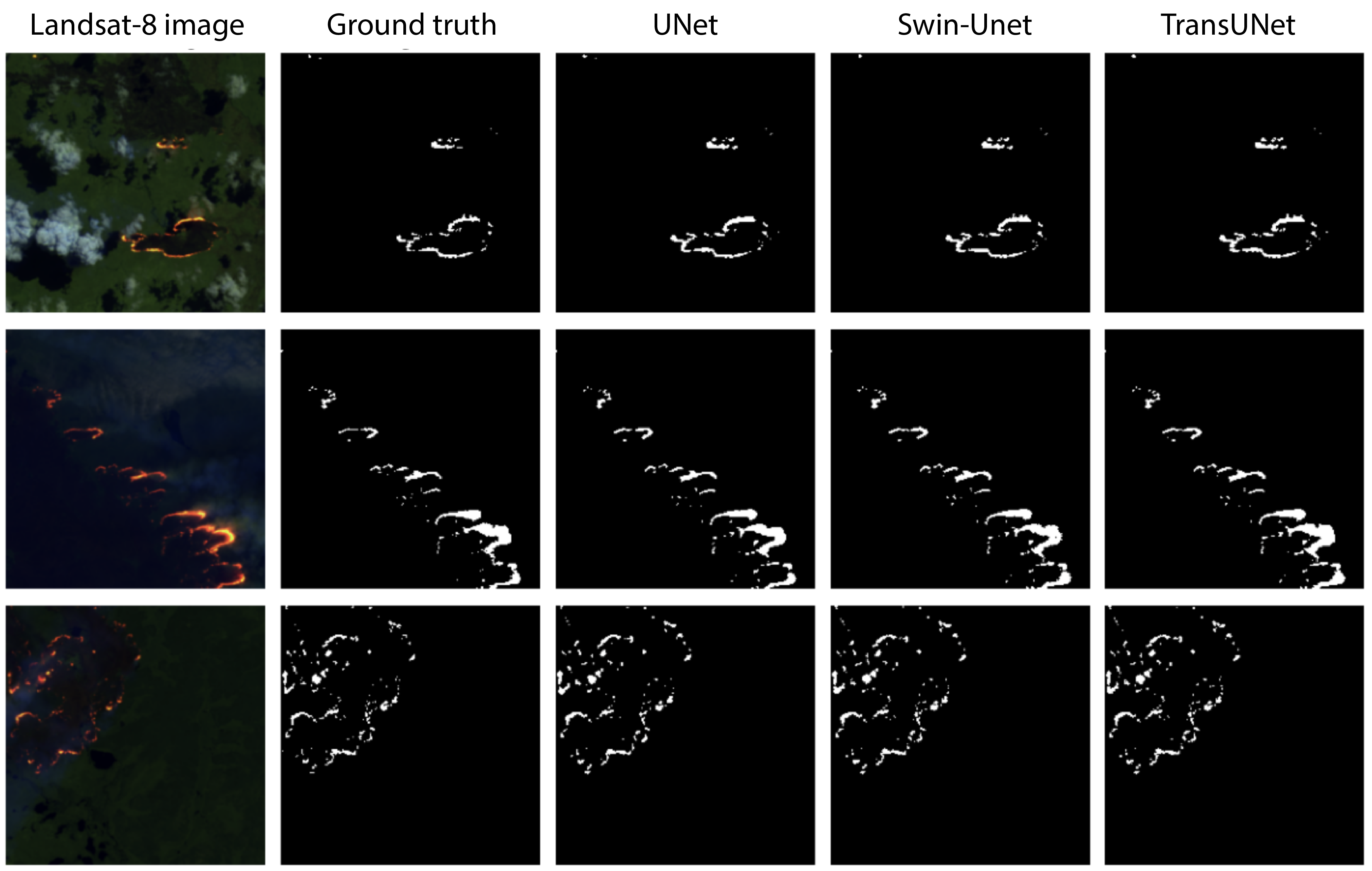}
	\caption{Inference results on some of the test-set images for each model. Visually, the results look almost indistinguishable.}
	\label{fig:results}
\end{figure*}

\begin{table}
  \centering
  \begin{tabular}{@{}lcccc@{}}
    \toprule
    Method & Precision & Recall & F-score & IoU \\
    \midrule
    U-Net (10c) \cite{Pereira2021}  & 92.90 & 95.50 & {\bf 94.20} & 89.00 \\
    U-Net (3c) \cite{Pereira2021}  & 91.90 & 95.30 & 93.60 & 87.90 \\
    U-Net-Light (3c) \cite{Pereira2021}  & 90.20 & {\bf 96.50} & 93.20 & 87.30 \\
    \midrule
    TransUNet \cite{chen2021transunet} & 88.46 & 86.88 & 87.66 & 87.49 \\
    Swin-Unet \cite{liu2021swin} & 88.28 & 92.30 & 90.24 & 89.93 \\
    Our UNet & {\bf 93.37} & 93.96 & 93.67 & {\bf 93.58} \\
    \bottomrule
  \end{tabular}
  \caption{Comparison of model performance on different metrics. Two of our models, UNet and Swin-Unet, outperforms the baseline models from Pereira et al.}
  \label{tab:table}
\end{table}

\section{Results \& Discussion}

Summary of model performance against different metrics is shown in Table \ref{tab:table}. The evaluation was done on four metrics: Precision, Recall, F-score, and Intersection over Union (IoU). Precision is an indication of how many correct fire pixels our algorithms detect out of all the detected fire pixels, recall is the measure of correct fire pixels our algorithms detect out of all the ground truth fire pixels, and F-score combines both precision and recall by taking a harmonic mean. On the other hand, IoU checks for the overlap between predicted and ground truth masks and is a common metric for object detection and segmentation problems. 

We were able to beat the models introduced by Pereira et al. \cite{Pereira2021} with both our UNet as judged by IoU and precision and with our Swin-Unet just in IoU. Our Swin-Unet drove a 0.93\% percent increase in IoU, while our UNet increased IoU by 4.58\%. Additionally, the improvement of precision by our UNet model was 0.47\% as compared to the precision of the best model presented by Pereira et al. A comparison of results produced by our models on the test set is shown in Figure \ref{fig:results}. 

Better performance of the ViT based models was expected as compared to the CNN models, yet our results demonstrate that the CNN based models achieved the best performance. A likely explanation as to why our ViT based models, particularly our Swin-Unet model, under performed relative to the UNet models is because the pre-trained weights of our model were optimized for the classification of everyday objects such as people, cars, buildings, trees, etc from RGB images. This is a problem because, first, our images were in the form of three spectral channels rather than RGB. Second, remote sensing imagery operates at a significantly larger scale than that of hand held camera images. Implementing pre-trained weights more appropriate for the regional to global scale of satellite imagery will likely increase our Swin-Unet model performance to a point comparable to our UNet architecture and greater than that of the UNet models used by Pereira et al. \cite{Pereira2021}.

Coming to our UNet model, it contains two additional aspects that we feel contributed to its improved performance. The first was dilated convolutions around the model's bottleneck, and the second was strided convolutions, as recommended in \cite{agarwal2021dilated}. Strided convolutions help to keep inter-feature dependencies intact and prevent feature loss. Dilated convolutions, on the other hand, aid in expanding the network's receptive field. As a result, they will have a broader context, similar to what ViTs aspire to. We were unable to prove the specific cause of improvement via any ablation study due to time constraints.

\section{Conclusion}

We have provided a successful application of a state-of-the-art computer vision algorithm to wildfire detection with promising results. Utilizing a dataset established by Pereira et al., we expand upon their study by testing the performance of two ViT architectures, TransUNet and Swin-Unet, against our own UNet model. We find that Swin-Unet architecture outperformed the TransUNet architecture and yielded a higher IoU than each of the UNet models presented by Pereira et al. \cite{Pereira2021}. However, our UNet model performed better than every model in precision and IoU, indicating that CNN based models may still be more equipped for fire detection from satellite images. However, our results indicate that Swin-Unet architecture is a promising avenue to pursue because of its performance versus the UNet architectures and reduced computational complexity. In future studies, the performance of the Swin-Unet model can be enhanced by pre-training the transformer, which we hope will drive the model accuracy higher than our UNet model.



\begin{figure*}[t]
	\centering
	\includegraphics[width=0.9\linewidth]{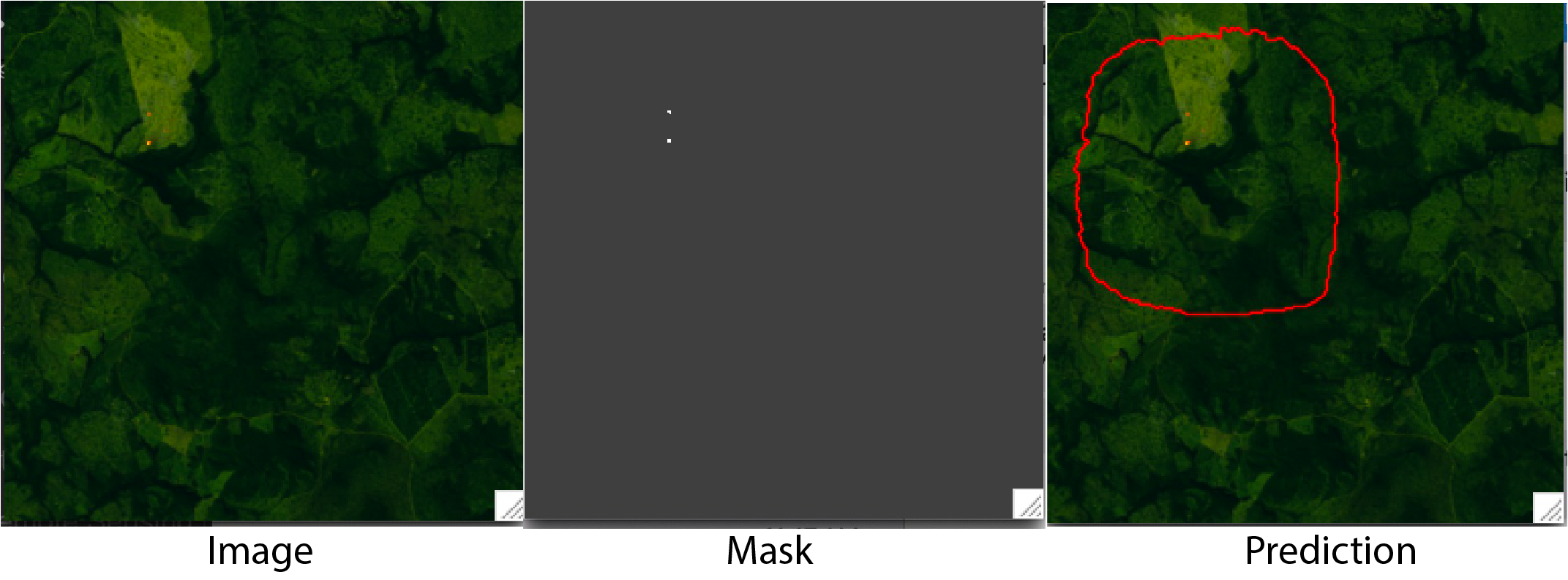}
	\caption{Inference result example for the Mask2Former instance detection model. This is one of its better results. Clearly, there are more favorable options.}
	\label{fig:maskformer}
\end{figure*}

\section*{Appendix}
In addition to the Swin-Unet and the TransUNet models, the Mask2Former model from the Facebook Research division \cite{chang21maskformer} was trained on the Pereira et al. 2021 dataset \cite{Pereira2021}. Mask2Former is a generalized architecture for image segmentation tasks (panoptic, instance, or semantic). Mask2Former uses a novel masked-attention technique that constrains cross-attention to within provided mask regions. The authors reduced training time by 3-fold and also beat the baseline for four popular image segmentation datasets. While this architecture is enticing, it suffers from confusing open-source software documentation and less-than happy maintainers, making understanding the vagueries of their in-house vision model NP-hard. Additionally, Mask2Former necessitates the inclusion of a bounding box on masked training data, which had to be done post-hoc for this dataset. In the end, IoU values hovered around 0.3\%, which is likely a result of improper training (Fig.  \ref{fig:maskformer}). As a result, these accuracy metrics are buried in the appendix.

\bibliographystyle{unsrtnat}
\bibliography{references}  






\end{document}